\documentclass[conference]{IEEEtran}
\IEEEoverridecommandlockouts
\usepackage{cite}
\usepackage{amsmath,amssymb,amsfonts}
\usepackage{graphicx}
\usepackage{textcomp}
\usepackage{xcolor}
\usepackage{nameref}
\usepackage{algorithm}
\usepackage{algpseudocode}
\usepackage{makecell}

\def\BibTeX{{\rm B\kern-.05em{\sc i\kern-.025em b}\kern-.08em
    T\kern-.1667em\lower.7ex\hbox{E}\kern-.125emX}}

\makeatletter
\newcommand{\linebreakand}{%
  \end{@IEEEauthorhalign}
  \hfill\mbox{}\par
  \mbox{}\hfill\begin{@IEEEauthorhalign}
}
\makeatother

\begin{document}

\title{
HaptiCharger: Robotic Charging of Electric Vehicles Based on Human Haptic Patterns
}

\author{

\IEEEauthorblockN{Oussama Alyounes\textsuperscript{\textsection}, 
Miguel Altamirano Cabrera\textsuperscript{\textsection} and 
Dzmitry Tsetserukou}
\IEEEauthorblockA{\textit{Skolkovo Institute of Science  
and Technology}\\
Moscow, Russia \\
oussama.alyounes@skoltech.ru, miguel.altamirano@skoltech.ru, d.tsetserukou@skoltech.ru }

}
\maketitle
\begingroup\renewcommand\thefootnote{\textsection}
\footnotetext{These authors contributed equally.}
\endgroup
\begin{abstract}

The growing demand for electric vehicles requires the development of automated car charging methods. At the moment, the process of charging an electric car is completely manual, and that requires physical effort to accomplish the task, which is not suitable for people with disabilities. Typically, the effort in the automation of the charging task research is focused on detecting the position and orientation of the socket, which resulted in a relatively high accuracy, $\pm$5 mm, and $\pm$10 degrees. However, this accuracy is not enough to complete the charging process. In this work, we focus on designing a novel methodology for robust robotic plug-in and plug-out based on human haptics to overcome the error in the orientation of the socket. Participants were invited to perform the charging task, and their cognitive capabilities were recognized by measuring the applied forces along with the movements of the charger. Eventually, an algorithm was developed based on the human's best strategies to be applied to a robotic arm.

\end{abstract}

\begin{IEEEkeywords}
Peg-in-hole, haptics, force-torque sensor.
\end{IEEEkeywords}

\section{Introduction}


Electric vehicles (EVs) are the future for zero-emission vehicles and the milestone of the ecological transportation unit \cite{AutomotiveWorld}. Over one million EVs were sold in 2017, and more than 10 million in 2022 \cite{iea}, with an expected increase in sales reaching 30\% by 2030 \cite{electricCarsSales}. In addition, it is expected that, by 2035, all new cars registered in Europe will be zero-emission \cite{zero_emission_2035}. While EVs are being developed focusing on their performance and cost-effectiveness, the operation of charging them is currently fully manual, with people required to connect the charging cable by themselves \cite{intro_technologies}. A survey of people with impairments was conducted by the British Research Institute for Disabled Consumers (RiDC) showing that only 25\% have the desire to use EVs \cite{RiDC}. The reason behind that is the unsuitable charging stations for them, which require physical activity to reach difficult places to plug in the charger. The survey also showed that this number increased to 61\% when the charging infrastructures were promised to be enhanced. The increased demand for EVs alongside the desire to meet consumers needs require the implementation and development of customer-friendly and novel charging infrastructure methods.

\begin{figure}[t!]
    \centering
    \includegraphics[width=0.47\textwidth]{ 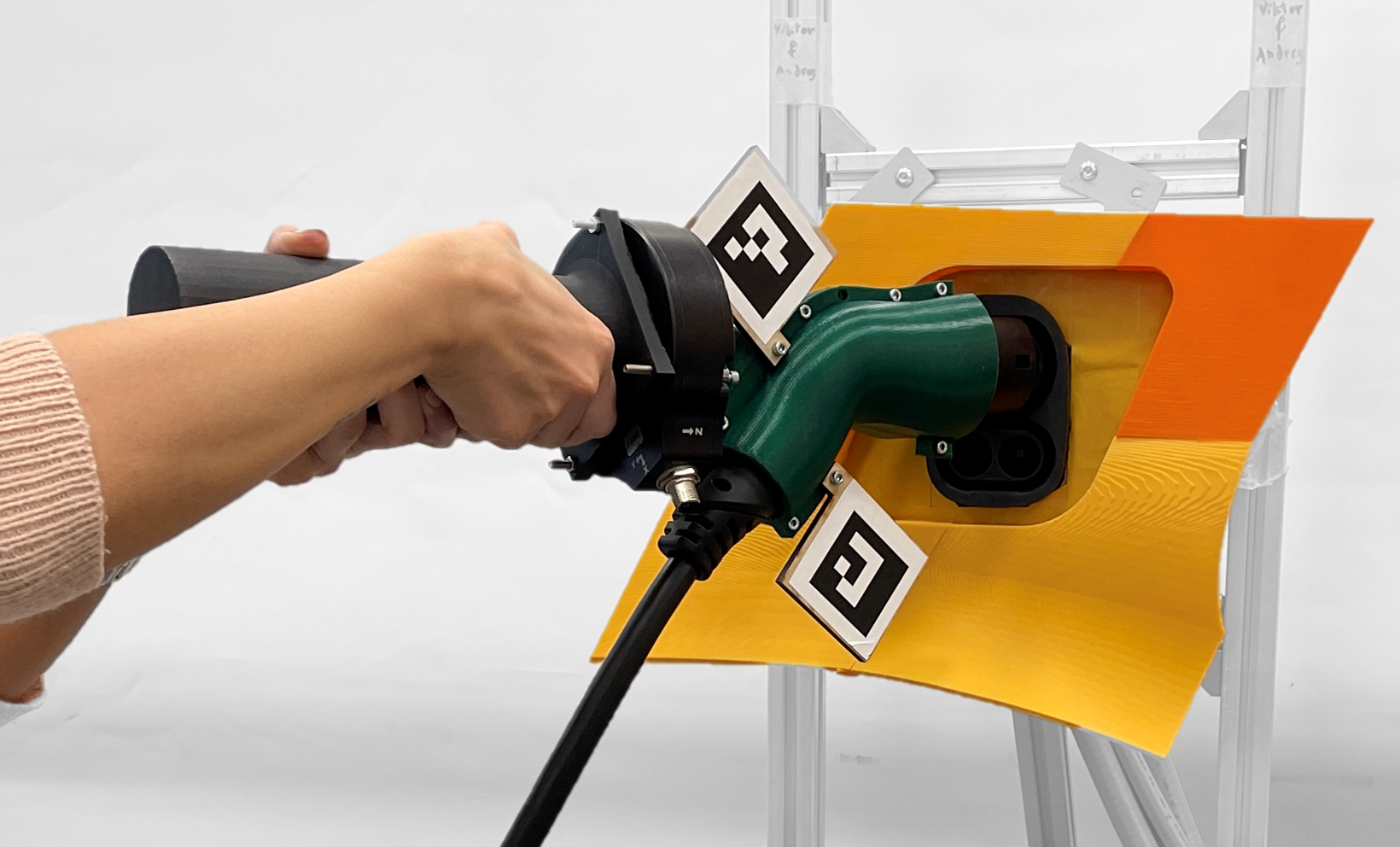}
    \vspace{-0,25 cm}
    \caption{Data acquisition experimental setup.}
    \label{fig:setup}
    \vspace{-0.35cm}
\end{figure}

The electric charging operation is an application of the peg-in-hole assembly task, which has been studied and applied in different practical tasks, such as large-scale component assembly \cite{Wan_large_scale} and screw plug-in \cite{screws_insertion}. Robotic assembly tasks require a high degree of repeatability and flexibility, which are usually achieved by hard-coding precise positions and trajectories to respond to the high accuracy required in the process \cite{HU2011715}. To increase the success rate, force-torque (FT) sensors were implemented in the process \cite{pih_example1}.
Human-inspired compliant was also designed with an FT sensor to detect the direction of the movement of the peg \cite{pih_example4}. However, human behavior was only studied by observation, with no attempt to record and process the data to achieve an optimal controller.

Humans use two main senses to carry out assembly tasks. The sight sense estimates the parts' locations and roughly matches them, and the kinesthetic sense finalizes the assembly by plugging the peg into the hole \cite{Abdullah2015}. Itabashi K. et al. suggested an approach for reproducing human skill in the peg-in-hole task that utilizes a sequence of impedance parameters identified from human demonstration \cite{Itabashi}.
By analyzing human motion conditions, it is possible to build manipulator control architectures based on human skills for different applications \cite{screwingSk}.

The full automation of the EV charging process incorporates three phases: the localization of the electric port, the plug-in of the electric charger in the socket, and the plug-out. The first phase is a computer-vision-related (CV) task that detects the position and orientation of the socket with relatively small errors \cite{Zou2019}. The small clearances between the socket and the peg, as well as the alignment of many pins with varying radii, are the main challenges of the plug-in process. M. Jokesch et al. analyzed the error in the localization of the socket and applied an impedance controller to compensate the error in 4 out of 6 degrees of freedom (DOF) depending on the measured forces at the tool center point (TCP) \cite{jokesch2016generic}. However, this study used the robot KUKA LWR iiwa 7 R800 with seven DOFs that allowed for the insertion of the charger depending on the compliance of the KUKA robot. X. Lv et al. proposed an automatic system guided by a camera and an FT sensor, reporting a 92\% success rate \cite{surf_template_match_FT}. However, the orientation of the socket was precisely known during the plug-in.

\begin{figure}
 \centering
 \includegraphics[width=0.34\textwidth]{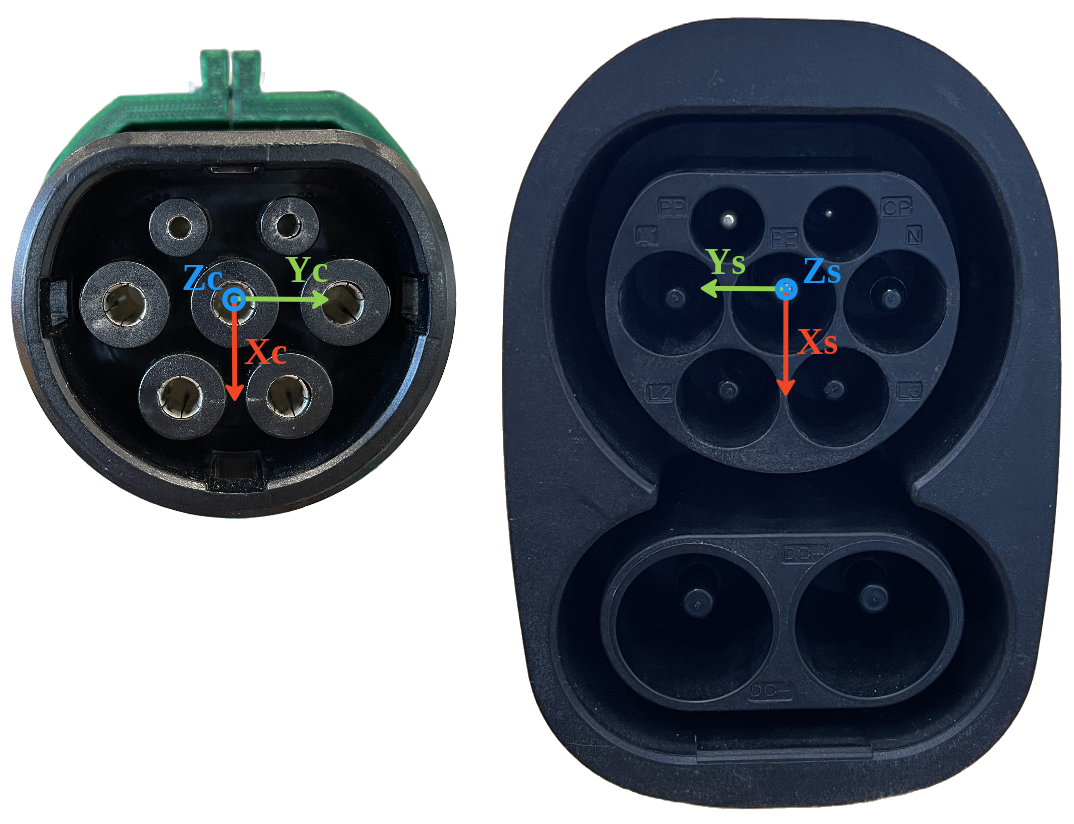}
 \caption{Socket and charger of type 2 used in the work.}
 \label{fig:socket&charger}
 \vspace{-0.35cm}
\end{figure}

This paper presents a study on recognizing the human haptics patterns implemented during the plug-in and plug-out phases of charging an electric car. An overview of the methodology and conditions of the experiment are presented in Section \ref{sec:Experiment_Design}. Two experiments were conducted, the first aims to identify the patterns that the users implement. The second experiment aims to analyze the human haptics patterns recognized in the first experiment and select the best strategy. In Section \ref{sec:A_Study_On_The_Spiral_Strategy}, an algorithm based on the best strategy is developed to be applied to a robotic manipulator.

\section{System Overview}

The general goal of this work is to develop an algorithm for plugging the charger in and out of the socket of an electric car based on human haptics. It is assumed that the first phase of localizing the socket is already done, and the focus of this work is compensating for orientation errors. An electric car charger and socket of type 2 were used in this study, as shown in Fig. \ref{fig:socket&charger}, where the plug and the socket are composed of seven different pins. The type of charger/socket had no influence on the work since the main focus of the paper is to investigate the human strategies to develop an algorithm.

The electric car charger was attached to a 6-DOF FT sensor 150-S from Robotiq, which works in the measuring ranges $\pm$150~N and $\pm$15~Nm for the forces and torques, and has an output rate of 100~Hz \cite{robotiq_sensor}. A charger holder was designed, 3D printed, and assembled with the charger to enable the user to carry the charger during the experiment. Two ArUco markers were attached to the charger setup to track its position. Furthermore, a camera of type Logitech HD 1080p C930e was mounted on a stand in a fixed position for the whole experiment to detect the positions of the markers. Two markers were used to cover in case one marker was not detected.

\begin{figure}[h]
 \centering
 \includegraphics[width=0.48\textwidth]{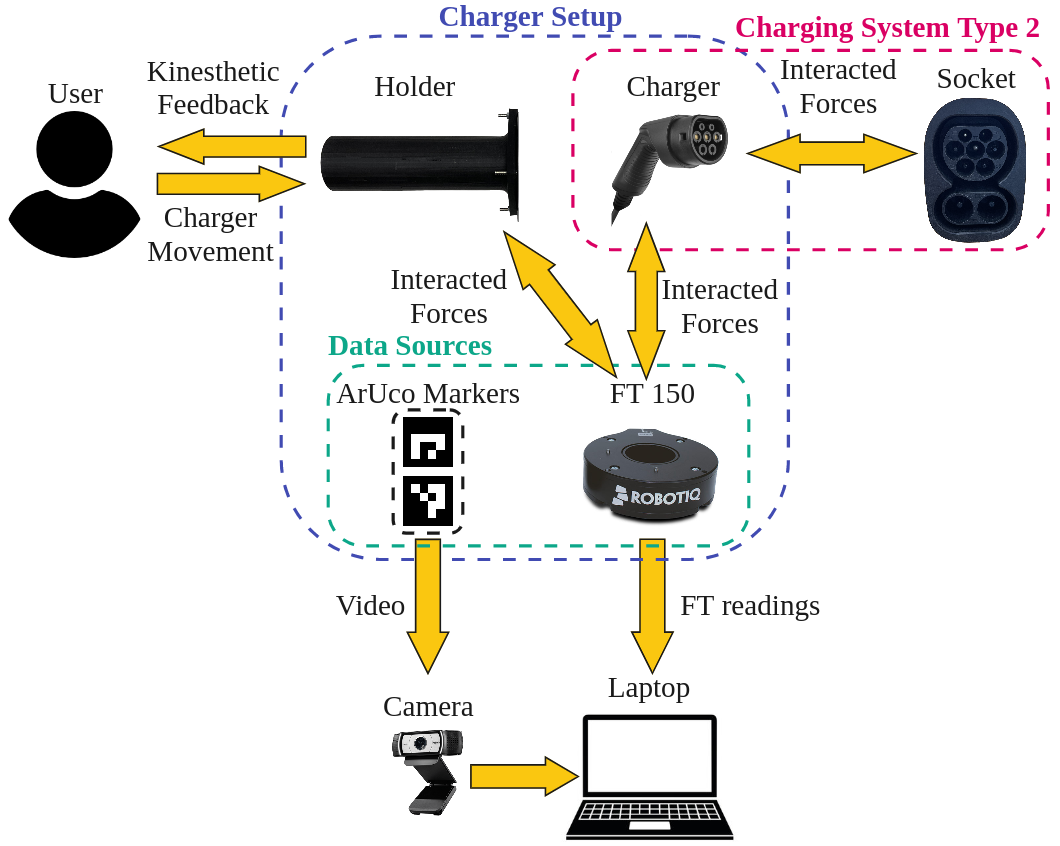}
 \caption{System architecture used for the human study experiment.}
 \label{fig:system_architecture_user}
 \vspace{-0.35cm}
\end{figure}

The system architecture for the data acquisition is depicted in Fig. \ref{fig:system_architecture_user}. This figure shows the interacted forces that are being transferred from the socket towards the user's hands, and the user is moving the charger to complete the task while readings from the FT sensor and ArUco markers are being recorded.

\section{Experimental Design}
\label{sec:Experiment_Design}
In this work, two experiments were conducted, the first experiment aimed to identify the haptic patterns that the users implement during the tasks, and the second experiment aimed to analyze these patterns recognized in the first experiment.

Several constraints were set during the experiments to have a unified context for all participants:

\begin{itemize}
  \item The standing position of the participants is in front of the socket base. This condition allows users to conduct the experiment without the need to walk. 
  
  \item Carrying the charger setup should be done only from the holder in obtain accurate readings from the FT sensor. .
  
  \item Each user performs the experiment blindly to maximize their dependence on the natural kinesthetic feedback.

  \item The starting point of the charger is at a close position to the socket. This allows the participants to start the plug-in phase without the need to search for the socket.
  
  \item The initial orientation of the charger is randomly tilted from the socket, with a total error of less than 10 deg.

  \item In the event of one participant fails to plug in the charger, the participant is assisted vocally by the experimenter.

  \item In the case of a complete failure by a participant, the participant was asked to repeat the task.

  \item Participants were told when they finished the plug-in phase in order to start with the plug-out phase.
  
\end{itemize}

Before the experiments, a training session was performed, where participants were explained the purpose of the experiment and the operation was demonstrated.

\textbf{Participants:} Twenty-three participants, seven females and sixteen males, capable of performing the experiment, aged 24.5$\pm$2.4 years, volunteered to participate in the experiments.

\begin{figure}[h!]
 \centering
 \includegraphics[width=0.39\textwidth]{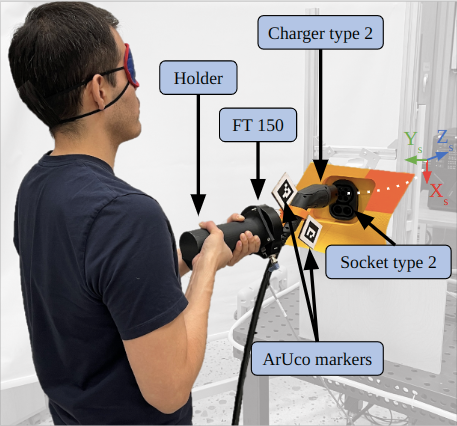}
 \caption{Data acquisition experimental setup.}
 \label{fig:experiment_setup}
 \vspace{-0.35cm}
\end{figure}

The setup for this experiment is depicted in Fig. \ref{fig:experiment_setup}, where one participant attempts to plug the charger into the socket.

\section{Strategies Identification Experiment}
\label{sec:Strategies_Identification}

The aim of the first experiment is to identify the natural strategies that users apply to accomplish the task. Participants were asked to perform the plug-in and plug-out tasks three times applying their intuitive strategy.

\subsection{Data Analysis}

\begin{figure}[]
 \centering
 \includegraphics[width=0.48\textwidth]{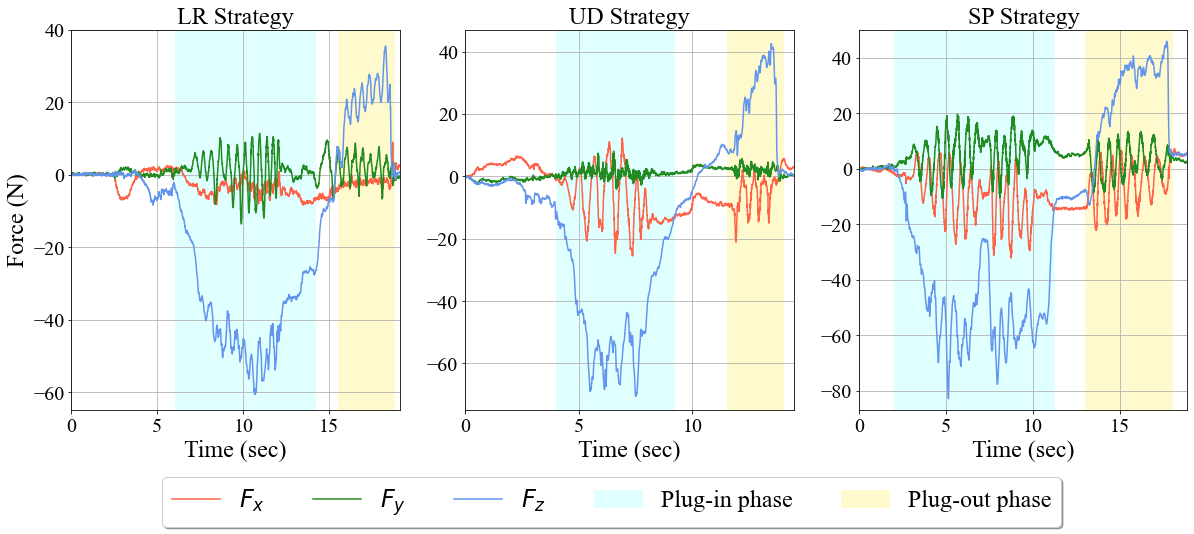}
 \caption{Forces changes for the three strategies applied by the participants.}
 \label{fig:user_force_3_strategies}
 \vspace{-0.35cm}
\end{figure}

\begin{figure}[]
 \centering
 \includegraphics[width=0.48\textwidth]{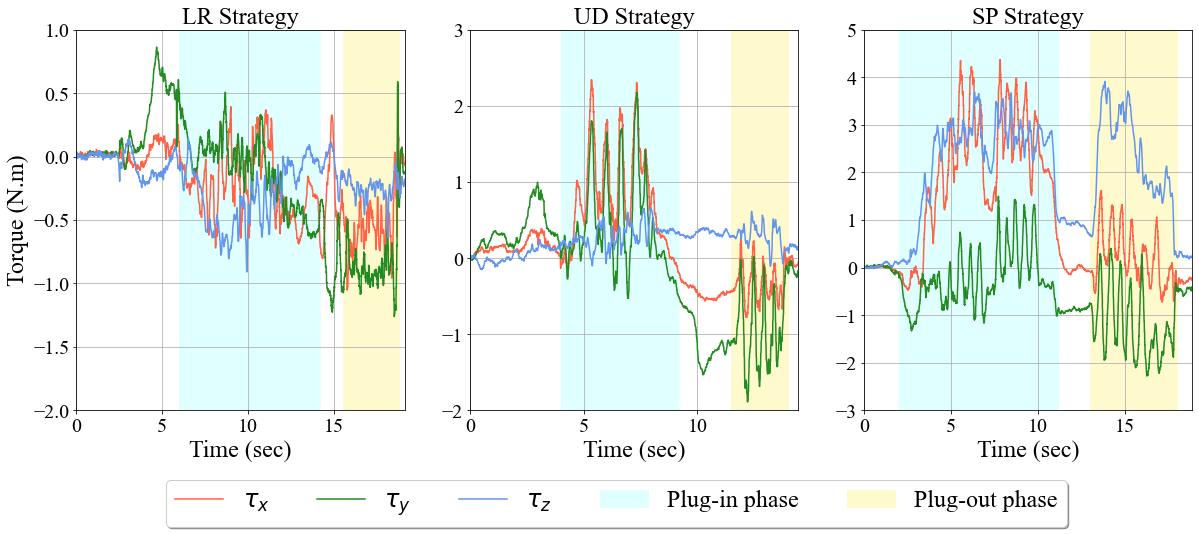}
 \caption{Torques changes for the three strategies applied by the participants.}
 \label{fig:user_torque_3_strategies}
 \vspace{-0.35cm}
\end{figure}

As a result of the first experiment, it was noted that only seventeen participants out of twenty three ($74 \%$) managed to accomplish the task with no assistance from the experimenter. The FT sensor readings showed that three strategies were applied by the participants and the changes in the FT values are shown in Fig.~\ref{fig:user_force_3_strategies} and Fig.~\ref{fig:user_torque_3_strategies}.

The first strategy mainly affects $F_y$ (and $\tau_x$) which indicates a movement in the left-right directions. The second strategy affects $F_x$ (and $\tau_y$) which indicates a movement in the up-down directions. And the third one affects both $F_x$ and $F_y$ ($\tau_x$, $\tau_y$) which indicates a movement in all directions. All strategies affect $F_z$ with negative values during the plug-in and positive values during the plug-out. 

\begin{figure*}[t]
 \centering
 \includegraphics[width=0.95\textwidth]{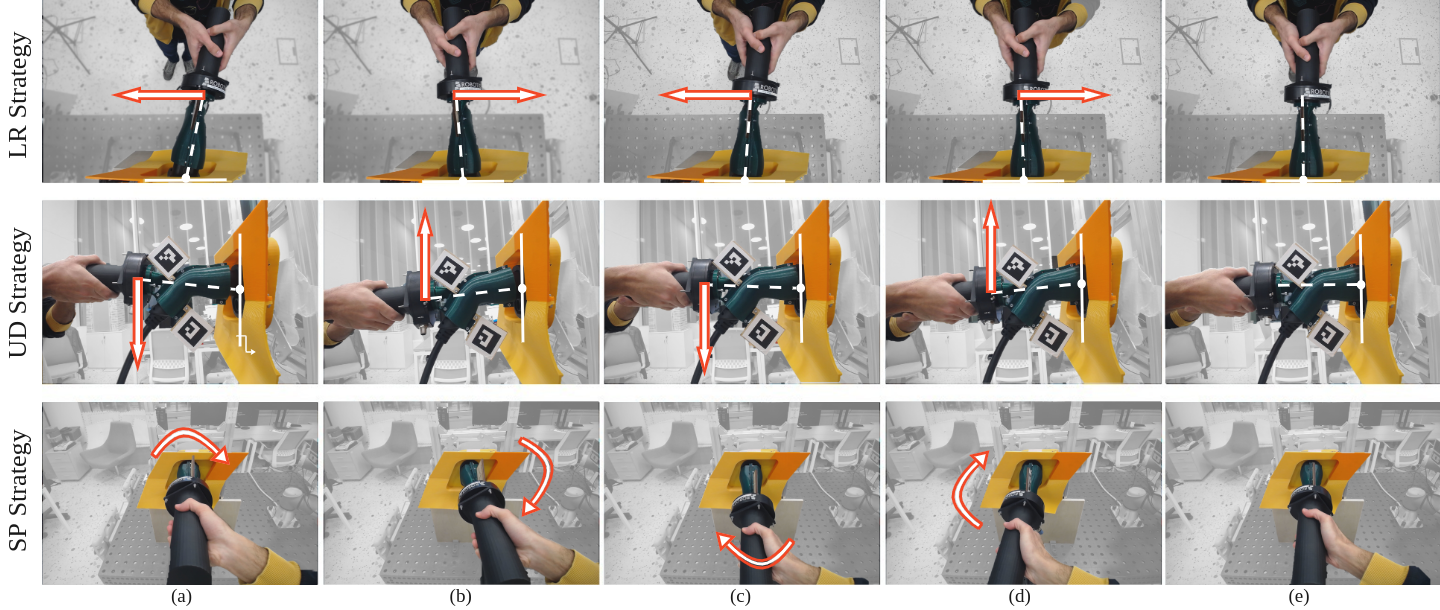}
 \caption{Strategies applied by participants to perform the plug-in of the charger inside the socket. LR strategy, moving the charger in the left-right direction. UD Strategy, moving the charger in the up-down direction. SP strategy, moving the charger in a circular direction. Steps (a), (b), (c), (d) and (e) show the steps to accomplish the plug-in for each strategy.
The arrows shown on each step represent the direction of movement required to reach the next step.}
 \label{fig:different strategy of plug in}
 \vspace{-0.35cm}
\end{figure*}

We named the strategies after the movement of the charger: left-right (LR strategy), up-down (UD strategy), and spiral (SP strategy). Fig. \ref{fig:different strategy of plug in} shows the steps of each strategy. By following these steps from (a) to (e), a full plug-in is being achieved. The plug-out can be achieved by following the opposite order. 

It is important to point out that our statistics showed that during the plug-in phase, 6 participants ($26.1 \%$), 2 participants ($8.7 \%$), and 15 ($65.2 \%$) participants depended on the LR, UD, and SP strategies, respectively. During the plug-out phase, 7 participants ($30.4 \%$), 5 participants ($21.8 \%$), and 2 participants ($8.7 \%$) depended on the LR, UD, and SP strategies, respectively. There were 9 participants ($39.1 \%$) who did not apply any strategy during the plug-out phase, but instead made a straight backward movement. This is explained because participants remembered the path they needed to follow and only moved the charger in the negative direction of $z_c$ axis.

It can be seen that the force readings are more stable and thus more valuable compared to the torques. In the next experiment, we will focus on the force readings.

\section{Strategies Comparison Experiment}
\label{sec:Strategies_Comparison}

After identifying the main strategies from the human behavior, a new experiment was conducted to compare these strategies regarding the ability to accomplish the task, the required time, and the applied forces. Participants were asked to repeat the task three times, applying one strategy at a time.

\subsection{Data Analysis}

\subsubsection{Task Completion}
It was found that, without any external help from the experimenter, 17 ($77 \%$), 20 ($91 \%$) and 24 ($100 \%$) of the participants finished the task using the LR, UD, and SP strategies, respectively. This resulted from the fact that LR and UD strategies do not correct errors in all orientations and this gives an advantage to the SP strategy over the others.

\subsubsection{Time Comparison}

\begin{figure}[]
 \centering
 \includegraphics[width=0.43\textwidth]{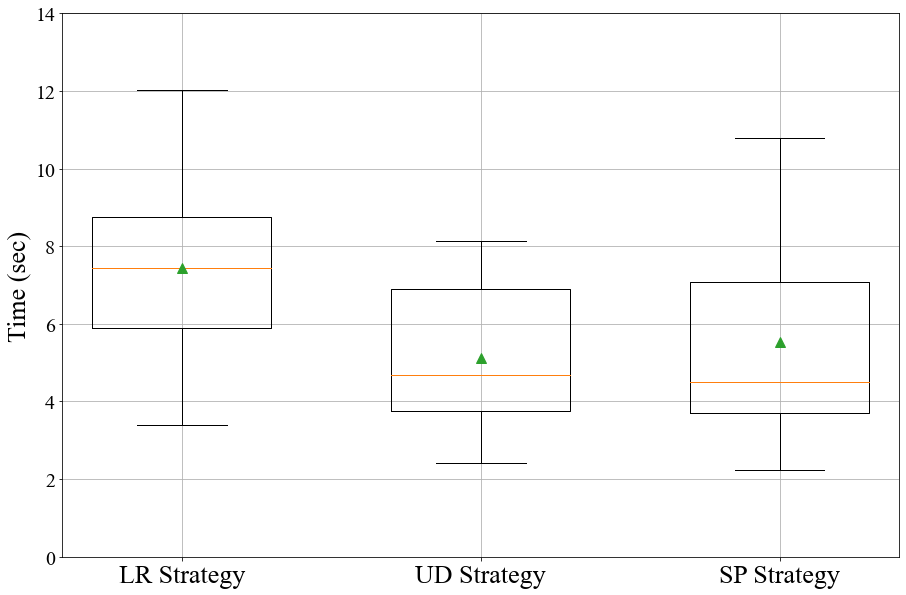}
 \caption{The time required by participants to plug in the charger depending on the different strategies.}
 \label{fig:time2plugin_second_experiment}
 \vspace{-0.35cm}
\end{figure}

The mean time required to plug in the charger was 7.43 sec, 5.12 sec and 5.52 sec for the LR, UD, and SP strategies, respectively. While the mean time required for the plug-out was 2.70 sec, 1.98 sec, 2.51 sec. Fig.~\ref{fig:time2plugin_second_experiment} 

shows the time required by participants to plug in the charger. 
A single-factor repeated-measures ANOVA with a significance level of $p<0.05$ was used to evaluate the statistical significance of the differences in the required time. ANOVA showed that there is a significant difference in the plug-in time among the strategies ($F (2,42) = 4.5554, \: p = 0.0162 < 0.05$). To know which strategies caused this difference, we made a pairwise comparison by running a t-test. It was found that there is a significant difference between the LR and SP strategies ($p = 0.0376 < 0.05$) and between the LR and UD strategies ($p = 0.0068 < 0.05$). However, there is no significant difference between the SP and UD strategies ($p = 0.6170 > 0.05$). For the plug-out time, there was no significant difference among the three strategies ($F (2,42) = 1.1908, \: p = 0.3140$).

\subsubsection{Force Comparison}
 
Table~\ref{table:min and max forces from participants} shows the mean minimum and the mean maximum forces applied by participants for all strategies. We can notice from this table how the LR strategy has big absolute values for $F_y$, the UD strategy for $F_x$ and the SP strategy for both $F_x$ and $F_y$. For the $F_z$ values, all strategies affect this force in almost the same way.

\begin{table}[]
\centering
\caption{average minimum and maximum forces applied by participants during the plug-in and out phases measured in Newton.}
\label{table:min and max forces from participants}
\setlength{\tabcolsep}{4pt}
\renewcommand{\arraystretch}{1}
\begin{tabular}{ccccccc}
\Xhline{4\arrayrulewidth}
& & & & & & \\[-2ex]

 \textbf{Strategy} & \multicolumn{1}{c}{$\overline{min F_x }$}  & \multicolumn{1}{c}{$\overline{min F_y }$}  & \multicolumn{1}{c}{$\overline{min F_z }$} & \multicolumn{1}{c}{$\overline{max F_x }$}  & \multicolumn{1}{c}{$\overline{max F_y }$}  & \multicolumn{1}{c}{$\overline{max F_z }$} \\ [0.2ex] \Xhline{4\arrayrulewidth} 

\textbf{LR}   & -15.7                                              & -16.9                                         & -90.4           & 4.9                                           & 15.0                                          & 46.8                               \\ 
\textbf{UD}      & -21.8                                           & -7.0                                           & -84.4            & 12.3                                   & 7.4                                    & 51.3                             \\ 
\textbf{SP}       & -21.6                                          & -15.6                                        & -81.6             & 11.8                                  & 16.5                                    & 53.9                               \\ \Xhline{4\arrayrulewidth}

\end{tabular}
\vspace{-0.35cm}
\end{table}

The analysis of the three strategies regarding the time and the ability to accomplish the task showed that the SP strategy is the best strategy among all the strategies.

\section{A Study on the SP Strategy}
\label{sec:A_Study_On_The_Spiral_Strategy}

After choosing the SP strategy as the best strategy applied by the participants, a detailed study to exploit this strategy was conducted with the goal of developing an algorithm for the plug-in and plug-out phases using a robotic arm.

\subsection{Force Analysis}
\begin{figure}[h!]
 \centering
 \includegraphics[width=0.48\textwidth]{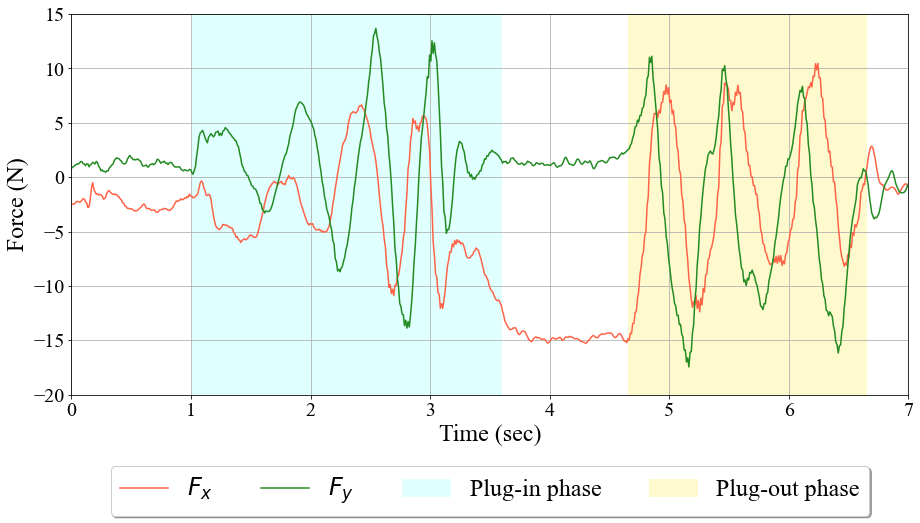}
 \caption{Changes of $F_x$ and $F_y$ for the plug-in and plug-out phases applying SP strategy from one participant.}
 \label{fig:human_force_xy_SP_strategy}
 \vspace{-0.35cm}
\end{figure}

\begin{figure}[h!]
 \centering
 \includegraphics[width=0.48\textwidth]{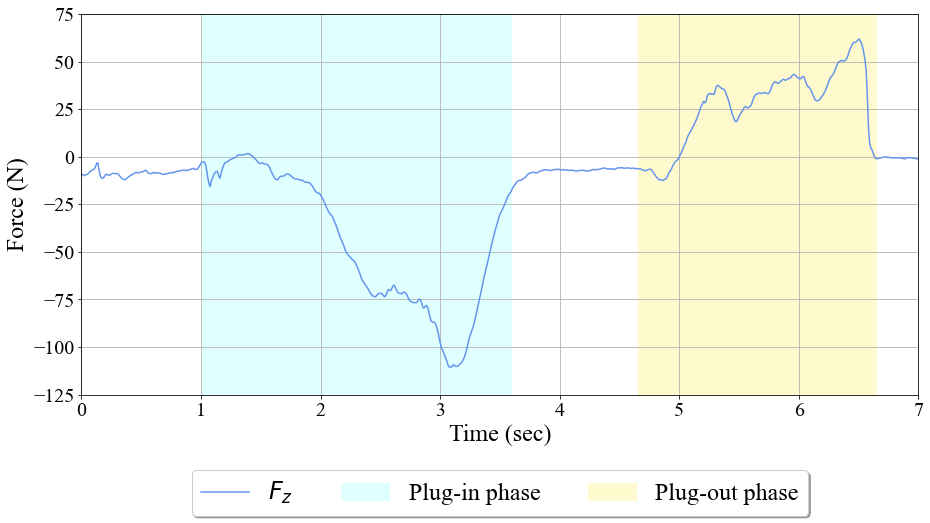}
 \caption{Changes of $F_z$ for the plug-in and plug-out phases applying SP strategy from one participant.}
 \label{fig:human_force_z_SP_strategy}
 \vspace{-0.35cm}
\end{figure}

SP strategy changes the forces, $F_x$ and $F_y$, in a shape of two sinusoidal waves with a difference in phase of $\phi = 90 \: degrees$. To analyze these changes, we measured the differences between two consecutive maximum and minimum values of the sinusoidal wave and took the maximum of these differences as $\Delta F_{x\_max}$ and $\Delta F_{y\_max}$. Fig. \ref{fig:human_force_xy_SP_strategy} shows the changes of $F_x$ and $F_y$ for one participant. This participant applied five waves in the plug-in and four waves in the plug-out. The required values are $\Delta F_{y\_max} = 27.5 \: N$ and $\Delta F_{x\_max} = 23.2 \: N$. From Fig. \ref{fig:human_force_z_SP_strategy}, we can also see the changes on $F_z$ for the same participant that were $F_{z\_plug-in} = -109.6 \: N$ and $F_{z\_plug-out} = 60.3 \: N$. 

\subsection{Movement Analysis}
Since the measured forces were the result of the interaction between the charger and the socket, we write the transformation from the charger to the socket $^{s}_{ch}T$. Using the camera we obtained the transformation from one ArUco marker to the camera $^{c}_{a}T$. We define the transformation from one marker to the charger $^{a}_{ch}T$. The camera and the socket were in fixed positions during the experiment, we wrote the transformation from the camera to the socket $^{s}_{c}T$. Eventually, we get:
$$^{s}_{ch}T = ^{s}_{c}T \: ^{c}_{a}T \: ^{a}_{ch}T$$

\begin{figure}[h!]
 \centering
 \includegraphics[width=0.12\textwidth]{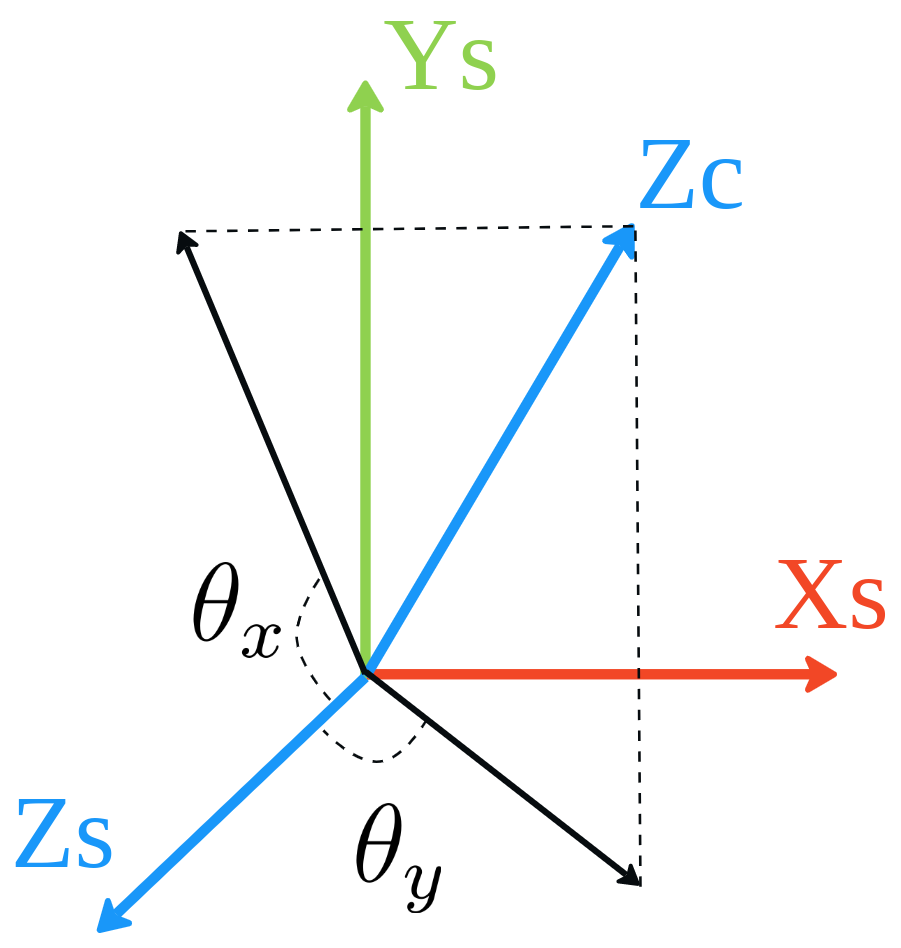}
 \caption{The coordinate systems of the socket and the charger with the calculated $\theta_x$ and $\theta_y$.}
 \label{fig:theta_x_y_coordinates}
 
\end{figure}

To measure the error in the orientation, we measured the angle between the two $z$ axes of the charger and the socket projected on the $z_s-x_s$ and the $z_s-y_s$ planes. Fig. \ref{fig:theta_x_y_coordinates} shows the measured angles, where $\theta_x$ is the angle projected on the $z_s-y_s$ plane representing the rotation around $x_s$ axis, while $\theta_y$ is the projected angle on the $z_s-x_s$ plane representing the rotation around $y_s$ axis. We write $^s_{ch}T$ as:

\begin{equation}
^s_{ch}T=\left(
\begin{array}{ccc|c}
r_{1,1} & r_{1,2} & r_{1,3} & \\
r_{2,1} & r_{2,2} & r_{2,3} & ^st_{ch}\\
r_{3,1} & r_{3,2} & r_{3,3} & \\ 
\hline 
0 & 0 & 0 & 1 
\end{array}
\right),
\end{equation}
where $r_{i,j} $ values represent the rotation matrix between the charger and the socket, and $^st_{ch}$ is the translation vector that transfers the origin of the charger coordinate system to the origin of the socket.

We calculate $\theta_x$ and $\theta_y$ as follows:
\begin{equation}
    \begin{aligned}
    \theta_x = \arccos(\frac{r_{3,3}}{\sqrt{r_{2,3}^2 + r_{3,3}^2}}), 
    \theta_y = \arccos(\frac{r_{3,3}}{\sqrt{r_{1,3}^2 + r_{3,3}^2}})
    \end{aligned}
\end{equation}

\begin{figure}[]
 \centering
  \includegraphics[width=0.48\textwidth]{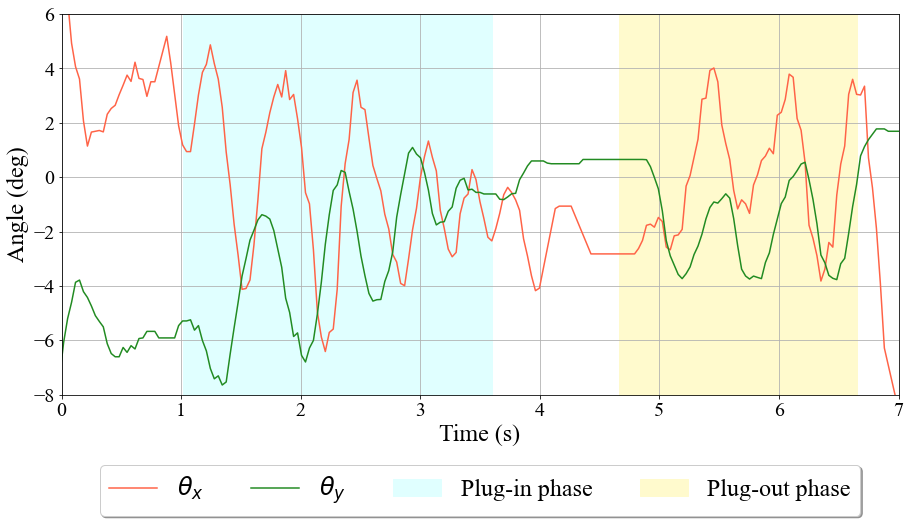}
 \caption{Changes of $\theta_x$ and $\theta_y$ for the plug-in and plug-out phases applying SP strategy from one participant.}
 \label{fig:angle-changes-user}
 \vspace{-0.35cm}
\end{figure}

Similar to the study on the forces, we measured the changes of angles ${\theta_{x\_ max}}$ and ${\theta_{y\_ max}}$. Fig. \ref{fig:angle-changes-user} shows the angle changes for one participant in the plug-in and plug-out phases which has the required values of $10.1$ and $6.9$ degrees, respectively.

The response time $t_{response}$ of the participants was also measured to determine the speed of the operation. The response time was measured as the time between two consecutive maximum and minimum values of the sinusoidal waves. 

We applied this study on the data from all participants and summarized the results in Table \ref{table: max,mean,std all data}.

\begin{table}[t]
\centering
\caption{Maximum, mean, and standard deviation of the angles and forces applied by the participants using spiral strategy.}
\label{table: max,mean,std all data}
\renewcommand{\arraystretch}{1.3}
\begin{tabular}{lcccc}
\Xhline{4\arrayrulewidth}

 \multicolumn{1}{l}{$\mathbf{X}$}& \multicolumn{1}{c}{$\mathbf{X_{mean}}$} &\multicolumn{1}{c}{$\mathbf{X_{max}}$} & \multicolumn{1}{c}{$\mathbf{\boldsymbol{\sigma_{X}}}$} & \multicolumn{1}{c}{$\mathbf{Unit}$} \\ \Xhline{4\arrayrulewidth}
$\Delta\theta_x$ &  $9.5 $                               & $14.8$                              & $2.1 $   & $ deg$                                  \\ 
$\Delta\theta_y $ & $6.8 $                               & $11.3$                               & $1.8 $ & $deg$                                                               \\ 
$\Delta F_x $& $27.7 $                               & $49.8$                              & $10.3 $ & $N$                                                              \\ 
$\Delta F_y $ & $32.6$                               & $47.1$                               & $7.9 $ & $N$                                      \\ 
$F_{z\_ plug-in} $ & $-81.6 $                               & $-103.7 (min)$                            & $14.5 $ & $N$                                                                 \\ 
$F_{z\_ plug-out}$ & $ 75.6$                              & $90.1$                               & $8.6 $ & $N$                                   \\ 

$t_{response} $ &    $0.26$                            & $0.37$                               & $0.08 $ & $sec$                                                              \\ \Xhline{4\arrayrulewidth}

\end{tabular}
\end{table}

\subsection{Plug-in Algorithm}

We visualize the relationship between the angle and the applied force as a second-order transfer function in the Laplace domain written following the notation in Eq. \ref{eq: TF second order}.

\begin{equation}
\label{eq: TF second order}
\frac{\theta(s)}{F(s)} = \frac{K_w \omega_n^2}{s^2 + 2\omega_n \zeta s + \omega_n^2},
\end{equation}
where $K_w, \omega_n, \zeta$ are the gain, natural frequency of the oscillation, and damping factor, respectively. These constants should be calculated depending on the human haptics study. For the charger of type 2 that was used in this study, the constants of the second-ordered system can be calculated from Table. \ref{table: max,mean,std all data}.

An algorithm based on the human haptics pattern was developed in Algo. \ref{alg:plugin}. The response of the system was chosen to be without overshoot ($\zeta = 1$), the settle time equal to the human speed ($t_s = t_{response}$), and the natural frequency is calculated based on the characteristics of a second-ordered system through the following formula \cite{transfer_function}:
$$t_s = \frac{4}{\zeta \omega_n}$$

\begin{algorithm}
\caption{An algorithm for the plug-in phase}
\label{alg:plugin}
\begin{algorithmic}[1]
\State $t_s \gets t_{response}$
\State $\zeta \gets 1$
\State $\omega_n \gets \frac{4}{\zeta t_s}$
\State $K_{lr} \gets \frac{\Delta F_x}{\Delta \theta_y}$
\State $K_{ud} \gets \frac{\Delta F_y}{\Delta \theta_x}$
\State $v_z \gets v_{const}$

\State Read $F_x, F_y, F_z$
\While{$F_z>F_{z\_plugin}$}
        \State solve $\Ddot{\theta_y} + 2\omega_n \zeta \dot{\theta_y} + \omega_n^2 \theta_y = K_{lr} \omega_n^2 F_{x}$ for $\theta_y$

        \State solve $\Ddot{\theta_x} + 2\omega_n \zeta \dot{\theta_x} + \omega_n^2 \theta_x = K_{ud} \omega_n^2 F_{y}$ for $\theta_x$

        \State send $\theta_x, \theta_y, v_z$

        \State $\theta_{x\_old} \gets \theta_x$
        \State $\theta_{y\_old} \gets \theta_y$
\EndWhile

\State $\theta_x \gets \theta_{x\_old}$
\State $\theta_y \gets \theta_{y\_old}$
\State $v_z \gets 0 $

\end{algorithmic}
\end{algorithm}

\section{Conclusion and Future Work}
In this paper, a study on the human behavior to accomplish the plug-in and plug-out phases of charging an electric car was conducted. Twenty-three participants volunteered for the study to carry out two experiments to study their behavior. The aim of the first experiment was to identify the main strategies applied by the participants, and it was conducted by asking the participants to perform the plug-in and plug-out phases without any restriction on the applied strategy.

The analysis of forces and torques showed that three strategies were intuitively applied by participants: left-right (LR strategy), up-down (UD strategy), and spiral (SP strategy). 

In the second experiment participant were asked to repeat the task three times applying one of the strategies each time. It was found that the SP strategy is the best strategy since it has a short time and is the only one that allowed the participants to finish the task with no additional help from the experimenter. Following this finding, an in-depth study was conducted for the SP strategy to analyze the response time, the forces, and the movements of the charger. Eventually, an algorithm to plug the charger in and out was developed to be applied to a robotic manipulator.

In the future, we aim to test our algorithm on a robotic charger. We will design a controller to automate the process of charging an electric car that takes advantage of the data driven from the SP strategy. A proper control method would be impedance control since it corrects the error between the environment (socket) and the manipulator (charger). Additionally, more investigation regarding the three studied strategies will be performed, and it will be applied on different applications.

\addtolength{\textheight}{-12cm}   






\begin{thebibliography}{xx}


\bibitem{AutomotiveWorld}
A. World, “Are electric vehicles still the future of automotive?” [Online]. Available: https://www.automotiveworld.com/articles/are-
electric-vehicles-still-the-future-of-automotive/

\bibitem{iea}
IEA. (2023) Electric car sales. [Online]. Available: https://www.iea.org/data-and-statistics/charts/electric-car-sales-2016-2023

\bibitem{electricCarsSales}
C. Guo and C. C. Chan, “Whole-system thinking, development control, key barriers and promotion mechanism for EV development,” Journal of Modern Power Systems and Clean Energy, vol. 3, no. 2, pp. 160–169, 2015.

\bibitem{zero_emission_2035}
E. Commission, “Zero emission vehicles: first ‘fit for 55’ deal will end the sale of new co2 emitting cars in europe by 2035.” [Online]. Available: https://ec.europa.eu/commission/presscorner/detail/en/ip 22 6462

\bibitem{intro_technologies}
B. Walzel, C. Sturm, J. Fabian, and M. Hirz, “Automated robot-based charging system for electric vehicles,” 03 2016.

\bibitem{RiDC}
RiDC, “Going electric?” [Online]. Available: https://www.ridc.org.uk/transport/going-electric

\bibitem{Wan_large_scale}
A. Wan, J. Xu, H. Chen, S. Zhang, and K. Chen, “Optimal path planning and control of assembly robots for hard-measuring easy-deformation assemblies,” IEEE/ASME Transactions on Mechatronics, vol. 22, no. 4, pp. 1600–1609, 2017.

\bibitem{screws_insertion}
Z. Liu, L. Song, Z. Hou, K. Chen, S. Liu, and J. Xu, “Screw insertion method in peg-in-hole assembly for axial friction reduction,” IEEE Access, vol. 7, pp. 148 313–148 325, 2019.

\bibitem{HU2011715}
S. Hu, J. Ko, L. Weyand, H. ElMaraghy, T. Lien, Y. Koren, H. Bley, G. Chryssolouris, N. Nasr, and M. Shpitalni, “Assembly system design and operations for product variety,” CIRP Annals, vol. 60, no. 2, pp. 715–733, 2011. [Online]. Available: https://www.sciencedirect.com/science/article/pii/S000785061100206X

\bibitem{pih_example1}
J. Takahashi, T. Fukukawa, and T. Fukuda, “Passive alignment principle for robotic assembly between a ring and a shaft with extremely narrow clearance,” IEEE/ASME Transactions on Mechatronics, vol. 21, no. 1, pp. 196–204, 2016.

\bibitem{pih_example4}
X. Li, R. Li, H. Qiao, C. Ma, and L. Li, “Human-inspired compliant strategy for peg-in-hole assembly using environmental constraint and coarse force information,” in 2017 IEEE/RSJ International Conference on Intelligent Robots and Systems (IROS), 2017, pp. 4743–4748.

\bibitem{Abdullah2015}
M. W. Abdullah, H. Roth, M. Weyrich, and J. Wahrburg, “An approach for peg-in-hole assembling using intuitive search algorithm based on human behavior and carried by sensors guided industrial robot,” IFAC- PapersOnLine, vol. 48, pp. 1476–1481, 2015.

\bibitem{Itabashi}
K. Itabashi, K. Hirana, T. Suzuki, S. Okuma, and F. Fujiwara, “Realization of the human skill in the peg-in-hole task using hybrid architecture,” in Proceedings. 1998 IEEE/RSJ International Conference on Intelligent Robots and Systems. Innovations in Theory, Practice and Applications (Cat. No.98CH36190), vol. 2, 1998, pp. 995–1000 vol.2.


\bibitem{screwingSk}
D. Mironov, M. Altamirano Cabrera, H. Zabihifar, A. Liviniuk, V. Liviniuk, and D. Tsetserukou, “Haptics of screwing and unscrewing for its application in smart factories for disassembly,” in Haptics: Science, Technology, and Applications. Cham: Springer International Publishing, 2018, pp. 428–439.

\bibitem{Zou2019}
P. Zou, Q. Zhu, J. Wu, and J. Jin, “An approach for peg-in-hole assembling based on force feedback control,” in 2019 Chinese Automation Congress (CAC), 2019, pp. 3269–3273.

\bibitem{jokesch2016generic}
M. Jokesch, J. Such`y, A. Winkler, A. Fross, and U. Thomas, “Generic algorithm for peg-in-hole assembly tasks for pin alignments with impedance controlled robots,” in Robot 2015: Second Iberian Robotics Conference. Springer, 2016, pp. 105–117.

\bibitem{surf_template_match_FT}
X. Lv, G. Chen, H. Hu, and Y. Lou, “A robotic charging scheme for electric vehicles based on monocular vision and force perception,” in 2019 IEEE International Conference on Robotics and Biomimetics (ROBIO). IEEE, 2019, pp. 2958–2963.

\bibitem{robotiq_sensor}
Robotiq, “Robotiq force torque sensor ft 150/300 instruction manual.” [Online]. Available: https://assets.robotiq.com/website-assets/support archives/document en/FT Sensor Instruction Manual
2016 11 18 PDF 20201105.pdf

\bibitem{transfer_function}
T. E. Fortmann and K. L. Hitz, An introduction to linear control systems. Crc Press, 1977.


\end{thebibliography}

%

\end{document}